\documentclass{isprs} % isprs class modified 23-04-2019 (Dennis Wittich)
\usepackage{subfigure}
\usepackage{multirow}
\usepackage{setspace}
\usepackage{amsfonts,amsmath}
\usepackage{geometry} % added 27-02-2014 Markus Englich
\usepackage{epstopdf}
\usepackage[labelsep=period]{caption}  % added 14-04-2016 Markus Englich - Recommendation by Sebastian Brocks
\usepackage[british]{babel} 
\usepackage[hang]{footmisc}
\usepackage{booktabs}
\usepackage{makecell}
\begin{document}

\title{A UNet Model for Accelerated Preprocessing of CRISM Hyperspectral Data for Mineral Identification on Mars}
\date{}

% KAO: Remove extra spacing

% Anonymous submissions, authors' names should not be visible
\author{
 Priyanka Kumari\textsuperscript{1}, Sampriti Soor\textsuperscript{2}, Amba Shetty\textsuperscript{3}, Archana M. Nair\textsuperscript{4}}
% \author{***** (for review, names must be rendered anonymous)}

% KAO: Remove extra newline
% Anonymous submissions, authors' affiliations should not be visible
\address{
	\textsuperscript{1 }IPDF, Department of Civil Engineering, Indian Institute of Technology Guwahati, India, 781039\\ - singh.priyanka854@gmail.com\\
	\textsuperscript{2 }IPDF, Center for Intelligent Cyber Physical System, Indian Institute of Technology Guwahati, India, 781039\\ - sampreetiworkid@gmail.com\\
	\textsuperscript{3 }Professor, Department of Water Resources and Ocean Engineering, National Institute of Technology Karnataka, India, 575025\\ – amba@nitk.edu.in\\
	\textsuperscript{4 }Associate Professor, Department of Civil Engineering, Indian Institute of Technology Guwahati, India, 781039\\ - nair.archana@iitg.ac.in
}
% \address{**** (for review, affiliations must be rendered anonymous)}

% If the corresponding author is NOT the final author, always add a % space before the subsequent comma, i.e.
% first author name\textsuperscript{a,}\thanks{Corresponding author} , % second author name \textsuperscript{b}, etc.
% thanks to Niclas Borlin 05-05-2016
% information on the corresponding author should not be used any longer and has been commented out
% C. Heipke, Jan 03,2024

% the use of the information of commissions and working groups should not be used any longer and has been commented out
% C. Heipke, Sept. 20,2022
%\commission{XX, }{YY} %This field is optional. If filled, XX and YY should be replaced by adequate numbers. See https://www2.isprs.org/commissions/
%\workinggroup{XX/YY} %This field is optional.
%\icwg{}   %This field is optional.

% KAO: Use times symbol
\abstract{
% Preprocessing hyperspectral data from the Compact Reconnaissance Imaging Spectrometer for Mars (CRISM) MTRDR dataset is critical for accurate mineral identification on the Martian surface. Traditional preprocessing steps, including spike removal, smoothing, wavelength standardization, scaling, and continuum removal, are computationally intensive, as they must be applied to each pixel across large datasets. This paper presents a neural network model based on the UNet architecture that integrates preprocessing directly within the model, eliminating the need for a separate pipeline. The proposed approach enhances spectral feature clarity and accelerates the mineral classification process without compromising accuracy. Comparative analysis demonstrates a processing time reduction of approximately 30\% and a classification accuracy increase of around 5\% over the traditional pipeline. Additionally, mineral mapping experiments highlight the model's effectiveness in identifying and mapping key minerals across different Martian terrains. This research offers a streamlined approach to preprocessing CRISM data, paving the way for real-time, large-scale mineral mapping on Mars.
Accurate mineral identification on the Martian surface is critical for understanding the planet's geological history. This paper presents a UNet-based autoencoder model for efficient spectral preprocessing of CRISM MTRDR hyperspectral data, addressing the limitations of traditional methods that are computationally intensive and time-consuming. The proposed model automates key preprocessing steps, such as smoothing and continuum removal, while preserving essential mineral absorption features. Trained on augmented spectra from the MICA spectral library, the model introduces realistic variability to simulate MTRDR data conditions. By integrating this framework, preprocessing time for an $800 \times 800$ MTRDR scene is reduced from 1.5 hours to just 5 minutes on an NVIDIA T1600 GPU. The preprocessed spectra are subsequently classified using MICAnet, a deep learning model for Martian mineral identification. Evaluation on labeled CRISM TRDR data demonstrates that the proposed approach achieves competitive accuracy while significantly enhancing preprocessing efficiency. This work highlights the potential of the UNet-based preprocessing framework to improve the speed and reliability of mineral mapping on Mars.
}


\maketitle

%\saythanks % added 28-02-2014 Markus Englich

%
%
% %% %% %% %% %% %% %% %% %% %% %% %% %% %% %% %% %% %% %% %% %% %% %% %%
% %% %% %% %% %%    I N T R O D U C T I O N    %% %% %% %% %% %% %% %% %%
% %% %% %% %% %% %% %% %% %% %% %% %% %% %% %% %% %% %% %% %% %% %% %% %%
%
%
\section{Introduction}\label{sec:introduction}
% KAO: Sloppy spacing ensures non-overfull lines. Can be removed if this is not an issue.
\sloppy
    Mineral identification on Mars is essential for understanding the planet's geological history, assessing past habitability, and identifying resources for future exploration and potential colonization \cite{kumari2023mineral}. The Compact Reconnaissance Imaging Spectrometer for Mars (CRISM) aboard the Mars Reconnaissance Orbiter captures hyperspectral data across visible to infrared wavelengths, enabling detailed analysis of Martian surface composition \cite{murchie2007compact}. After necessary atmospheric and photometric corrections, CRISM provides high-resolution spectral data, specifically the Map-Projected Targeted Reduced Data Record (MTRDR), which captures distinct spectral signatures of various minerals essential for accurate mineral mapping on the Martian surface \cite{viviano2014revised}.

The raw spectral data from MTRDR are inherently complex and require extensive preprocessing to transform them into a format suitable for mineral identification. This preprocessing addresses various challenges such as noise, baseline curvature, and distortions that obscure diagnostic absorption features critical for identifying minerals:
\begin{enumerate}
    \item \textit{Wavelength range selection and scaling:} Hyperspectral data from CRISM spans a wide range of wavelengths. However, not all wavelengths contribute meaningful information for mineral identification, and some may introduce noise or redundancy. Selecting an optimal wavelength range ensures that only the most informative bands are retained. Scaling these bands normalizes the data, ensuring consistency across different datasets, which is critical when comparing or combining spectra from various sources.
    \item \textit{Spike removal and smoothing:} CRISM data often contain random spikes and fluctuations due to instrument noise or external interference. These spikes can distort the spectral shape, leading to misinterpretation of mineralogical features. Spike removal techniques identify and eliminate these anomalies, while smoothing algorithms refine the spectral curve, reducing high-frequency noise while preserving the integrity of key absorption features.
    \item \textit{Continuum removal:} The spectral baseline, or continuum, represents the broad curvature of a spectrum caused by factors like albedo variations or systematic instrument effects. This baseline can mask subtle absorption features, which are crucial for mineral identification. Continuum removal eliminates this curvature by fitting and subtracting a continuum curve, leaving behind a spectrum that highlights distinct absorption bands. This step significantly enhances the visibility of mineralogical features, making it easier to match the processed spectrum with reference libraries like MICA \cite{saranathan2021adversarial}.
\end{enumerate}
These steps, typically part of a preprocessing pipeline \cite{kumari2023fully}, are crucial but computationally expensive, often taking over an hour for large MTRDR scenes on standard hardware. As data volume increases, the time required for traditional preprocessing becomes a significant bottleneck. 
Mineral classification on Mars is further complicated by the lack of ground-truth data, as field samples are unavailable. Instead, training data are derived from spectral libraries, with the MICA spectral library \cite{viviano2014revised} providing known signatures for expected Martian minerals. For realistic training, the MICA spectra are augmented to introduce noise and baseline curvature similar to real MTRDR data, preserving key absorption features while enhancing model robustness against natural data variability.

This study proposes a UNet-based neural network model \cite{ronneberger2015u} to streamline the preprocessing of MTRDR hyperspectral data. By integrating essential preprocessing steps within a single model, the proposed approach bypasses the need for a separate pipeline, significantly improving processing efficiency. The primary objectives are:
\begin{itemize}
    \item To replace traditional preprocessing pipelines with a neural network model capable of directly enhancing spectral clarity.
    \item To accelerate the preprocessing phase, reducing total time required from hours to minutes.
    \item To retain the accuracy of mineral classification by preserving critical spectral features essential for Martian mineral identification.
\end{itemize}

To assess its effectiveness, the model's preprocessed output is evaluated on MICAnet \cite{kumari2024micanet}, a mineral classification model, using labeled data from \cite{plebani2022machine}. This approach allows an indirect measure of model accuracy in identifying minerals across Martian locations. Compared to traditional preprocessing, the proposed UNet-based model reduces processing time by approximately 90\%, with significant reductions in pipeline complexity and achieves around 5\% enhancement in mineral identification accuracy.

% In summary, this paper introduces a UNet-based model that replaces conventional preprocessing pipelines, enabling rapid, accurate mineral identification on Mars. This advancement holds potential for large-scale mineral mapping, facilitating Martian surface analysis and contributing to broader planetary science and exploration objectives.

The structure of this paper is organized as follows. Section 1 introduces the study, covering its significance, related terminologies, and relevant past work in this domain. Section 2 presents the methodology, detailing data preparation in Section 2.1, the UNet-based model architecture in Section 2.2, and a comprehensive model specification and ablation study in Section 2.3. Section 2.4 provides a performance analysis, comparing the proposed model’s efficiency and accuracy against traditional preprocessing pipelines. Section 3 demonstrates mineral mapping outcomes on the Martian surface. Finally, Section 4 summarizes the key findings, offers concluding remarks, and suggests future directions for this research.

%
%
% %% %% %% %% %% %% %% %% %% %% %% %% %% %% %% %% %% %% %% %% %% %% %% %%
% %% %% %% %% %% %%  M E T H O D O L O G Y  %% %% %% %% %% %% %% %% %% %%
% %% %% %% %% %% %% %% %% %% %% %% %% %% %% %% %% %% %% %% %% %% %% %% %%
%
%
\section{Methedology}\label{sec:methodology}
% -- -- -- -- -- -- -- -- -- -- -- -- -- -- -- -- -- -- -- --
% -- -- -- -- --   Data Preparation   -- -- -- -- -- -- -- --
% -- -- -- -- -- -- -- -- -- -- -- -- -- -- -- -- -- -- -- --
\subsection{Data Preparation}\label{sec:dataprep}
    The training data for this study is generated from the MICA spectral library, which includes 31 mineral spectra across six major groups: iron oxides, primary silicates, ices, sulfates, phyllosilicates, carbonates, and hydrated silicates and halides. Table~\ref{tab:mineral_classes} provides the mineral classes and their corresponding groups. For testing the model, we use CRISM's Targeted Remote Sensing Data Record (TRDR), which provides high-resolution spectral data of the Martian surface, accessible via NASA PDS \cite{maki2004mer}. Authors in \cite{plebani2022machine} classified a substantial set of 592,413 TRDR spectra from over 70 images into 39 mineral categories, of which 28 match those in the MICA library. We randomly sample 200 spectra per label from this TRDR dataset, focusing on the wavelength range 1–2.6~$\mu$m from the available 1–3.47~$\mu$m, as most distinct mineral absorptions occur within this range, with data beyond it often exhibiting high noise levels.

\begin{table}[h!]
\centering
\caption{Mineral Classes and Groupings Considered in the MICA Spectral Library}
\label{tab:mineral_classes}
\begin{tabular}{ll}
\hline
\textbf{Mineral Groups} & \textbf{Mineral Classes}\\
\hline
\makecell[l]{Iron Oxides\\and\\Primary Silicates} & \makecell[l]{hematite, mg olivine, fe olivine,\\low calcium pyroxene,\\high calcium pyroxene} \\
\hline
Ices & CO\textsubscript{2} ice, H\textsubscript{2}O ice \\
\hline
Sulfates& \makecell[l]{alunite, poly hydrated sulfate ,\\ mono hydrated sulfate, bassanite, \\jarosite, gypsum, Fehydoxysulfate} \\
\hline
\makecell[l]{Phyllosilicates}& \makecell[l]{kaolinite, Al smectite,\\illite muscovite, Fe smectite, \\ Mg smectite, serpentine, chlorite }\\
\hline
Carbonates& Fe/Ca carbonate  carbon, Mg carbonate \\
\hline
\makecell[l]{Hydrated Silicates\\and Halides} & \makecell[l]{prehnite, epidote, hydrated silica,\\ analcime, chloride} \\
\hline
\end{tabular}
\end{table}

Since there is no labeled training data available to supervise mineral identification on the Martian surface, training data is synthesized from the MICA library through data augmentation. In natural CRISM MTRDR datasets, spectra within a single pixel often result from mixtures of multiple minerals. To align with this phenomenon, we create augmented data by generating synthetic spectra that simulate these mixtures.

To create an augmented spectrum, a random mineral \( m_1 \) is initially selected and assigned a random proportion \( p_1 \) in the range \([0, 1]\). For subsequent minerals, proportions are assigned based on the remaining proportion of 1. The mixing proportions for each mineral \( i \) can be defined as:
\[
p_i = \begin{cases} 
    U(0, 1) & \text{for } i=1 \\
    U(0, 1 - \sum_{j=1}^{i-1} p_j) & \text{for } i > 1 
\end{cases}
\]
where \( U(a, b) \) represents a uniform distribution between \( a \) and \( b \), and \( \sum_{i=1}^n p_i = 1 \).

Once proportions are determined, the mixed spectrum \( S_{\text{mix}}(\lambda) \) at each wavelength \( \lambda \) is calculated as:
\[
S_{\text{mix}}(\lambda) = \sum_{i=1}^n p_i \cdot S_i(\lambda)
\]
where \( S_i(\lambda) \) is the spectrum of mineral \( i \) at wavelength \( \lambda \).

To account for real data noise, Gaussian noise is added to each augmented spectrum:
\[
S_{\text{aug}}(\lambda) = S_{\text{mix}}(\lambda) + \mathcal{N}(0, \sigma)
\]
where \(\mathcal{N}(0, \sigma)\) represents Gaussian noise with mean 0 and standard deviation \(\sigma \in [0, 0.1]\), as higher values obscure the absorption features. The augmented spectra are also scaled within the selected wavelength range of 1 to 2.6 micrometer, as most minerals fall within this range, by setting the minimum value to 0 and the maximum to 1.

The generated dataset, therefore, closely represents the natural variability of mineral spectra in CRISM data, enabling the model to learn from realistic Martian surface spectral characteristics.

\begin{figure}
    \centering
    \includegraphics[width=\linewidth]{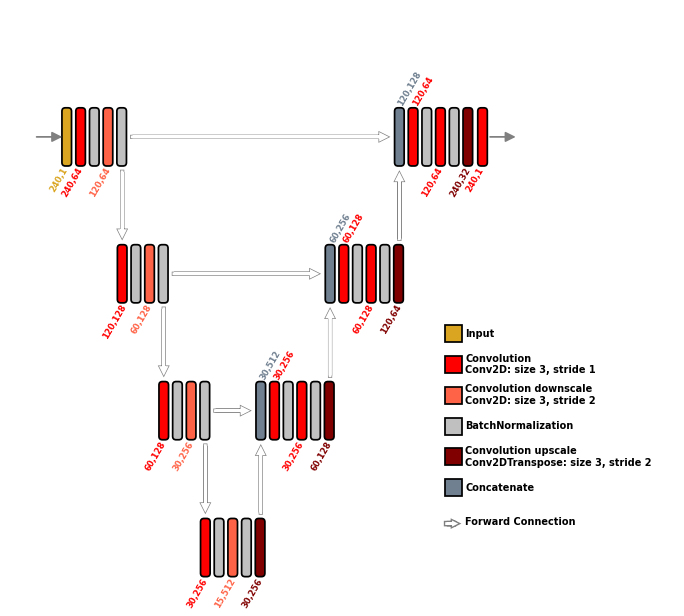}
    \caption{The U-net architecture used in this study. The architecture (IV-B) is fixed by an ablation study which is detailed in section \ref{sec:ablationstudy}.}
    \label{fig:unet}
\end{figure}
% -- -- -- -- -- -- -- -- -- -- -- -- -- -- -- -- -- -- -- --
% -- -- -- -- --  Model Architecture  -- -- -- -- -- -- -- --
% -- -- -- -- -- -- -- -- -- -- -- -- -- -- -- -- -- -- -- --
\subsection{Model Architecture}\label{sec:modelarchi}
    The proposed model architecture is a 1D UNet designed to preprocess CRISM MTRDR hyperspectral data, with 240 spectral bands within the 1 to 2.6 $\mu$m wavelength range, corresponding to the input dimension of $240 \times 1$. The UNet model includes an encoder-decoder structure with skip connections, allowing for detailed feature extraction and efficient upsampling.

\subsection*{Input Layer}
The input layer, $\mathbf{X}$, represents the hyperspectral data for each pixel, given by:
\begin{equation}
\mathbf{X} \in \mathbb{R}^{240 \times 1}
\end{equation}

\subsection*{Encoder Blocks}
Each encoder block comprises two convolutional layers. The first convolutional layer maintains the feature dimensionality while the second reduces the spatial resolution, thus downsampling the spectral information.

For the $i$-th encoder block, let $\mathbf{E}_i$ denote the feature map output:
\begin{equation}
\mathbf{E}_i = f(\mathbf{W}_i^{(1)} * \mathbf{E}_{i-1} + \mathbf{b}_i^{(1)})
\end{equation}
\begin{equation}
\mathbf{E}_i = f(\mathbf{W}_i^{(2)} * \mathbf{E}_i + \mathbf{b}_i^{(2)})
\end{equation}
where $\mathbf{W}_i^{(1)}$ and $\mathbf{W}_i^{(2)}$ are the weights for the first and second convolutional layers in the $i$-th encoder block, $\mathbf{b}_i^{(1)}$ and $\mathbf{b}_i^{(2)}$ are the corresponding biases, and $f$ denotes the activation function. Each convolutional layer is followed by a Batch Normalization layer, and the second layer has a stride of 2 to reduce dimensionality by half.

\subsection*{Bottleneck Block}
The bottleneck block connects the encoder and decoder sections. It consists of two convolutional layers followed by an upsampling layer to enable feature expansion for the decoder.

Let $\mathbf{B}$ denote the bottleneck output:
\begin{equation}
\mathbf{B} = f(\mathbf{W}_B^{(1)} * \mathbf{E}_N + \mathbf{b}_B^{(1)})
\end{equation}
\begin{equation}
\mathbf{B} = f(\mathbf{W}_B^{(2)} * \mathbf{B} + \mathbf{b}_B^{(2)})
\end{equation}
\begin{equation}
\mathbf{B} = f(\text{Upsample}(\mathbf{B}))
\end{equation}
where $\mathbf{W}_B^{(1)}$ and $\mathbf{W}_B^{(2)}$ are weights of the bottleneck block's convolutional layers, $\mathbf{b}_B^{(1)}$ and $\mathbf{b}_B^{(2)}$ are biases, and $\text{Upsample}$ denotes the Conv1DTranspose layer that doubles the feature map resolution.

\subsection*{Decoder Blocks}
The decoder blocks receive input from the bottleneck block and include concatenation layers for the skip connections from the encoder. Each decoder block has two convolutional layers that process the concatenated features, followed by an upsampling layer to increase resolution.

For the $j$-th decoder block, the feature map $\mathbf{D}_j$ is defined as:
\begin{equation}
\mathbf{D}_j = \text{Concat}(\mathbf{B}_{j-1}, \mathbf{E}_{N-j})
\end{equation}
\begin{equation}
\mathbf{D}_j = f(\mathbf{W}_j^{(1)} * \mathbf{D}_j + \mathbf{b}_j^{(1)})
\end{equation}
\begin{equation}
\mathbf{D}_j = f(\mathbf{W}_j^{(2)} * \mathbf{D}_j + \mathbf{b}_j^{(2)})
\end{equation}
\begin{equation}
\mathbf{D}_j = f(\text{Upsample}(\mathbf{D}_j))
\end{equation}
where $\text{Concat}$ denotes the concatenation operation for the skip connections, and $\text{Upsample}$ refers to the Conv1DTranspose operation.

\subsection*{Output Layer}
The output layer produces a final 1D representation with the same dimensionality as the input, ensuring compatibility with subsequent analysis tasks. The output, $\mathbf{Y}$, is expressed as:
\begin{equation}
\mathbf{Y} = \mathbf{W}_{\text{out}} * \mathbf{D}_1 + \mathbf{b}_{\text{out}}
\end{equation}
where $\mathbf{W}_{\text{out}}$ and $\mathbf{b}_{\text{out}}$ are the weights and biases of the output layer, respectively.

% \subsection*{Importance of the Architecture}
This architecture leverages the hierarchical feature extraction capabilities of the encoder-decoder structure. By using Conv1D layers, the model can capture spectral variations across wavelengths. The skip connections ensure that fine-grained spectral information is preserved across layers, critical for accurate mineral identification. This design balances feature reduction and preservation, optimizing computational efficiency and accuracy in mineral classification on the Martian surface.

% -- -- -- -- -- -- -- -- -- -- -- -- -- -- -- -- -- -- -- --
% -- -- -- -- --   Model Specifications   -- -- -- -- -- -- -- --
% -- -- -- -- -- -- -- -- -- -- -- -- -- -- -- -- -- -- -- --

\subsection{Model and Training Specifications}\label{sec:modelspec}
In our proposed 1D-UNet model, each convolutional layer is equipped with a Rectified Linear Unit (ReLU) activation function to introduce non-linearity while maintaining computational efficiency. The ReLU activation function is defined as:
\begin{equation}
f(x) = \max(0, x)
\end{equation}
which enhances the model’s capacity to learn complex representations in the spectral data.

\begin{figure}[!t]
    \centering
    \includegraphics[width=\linewidth]{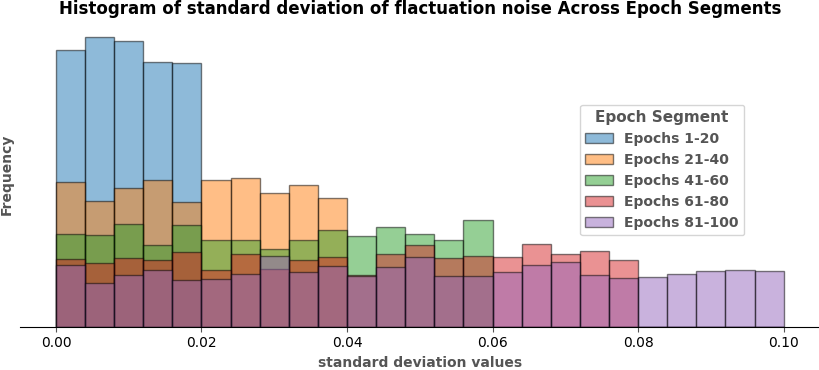}
    \par(a)\par\vspace{.5cm}
    \includegraphics[width=\linewidth]{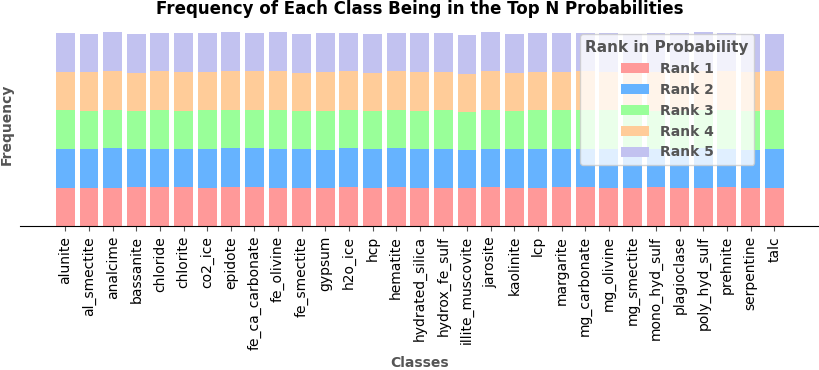}
    \par(b)\par\vspace{.5cm}
    \includegraphics[width=\linewidth]{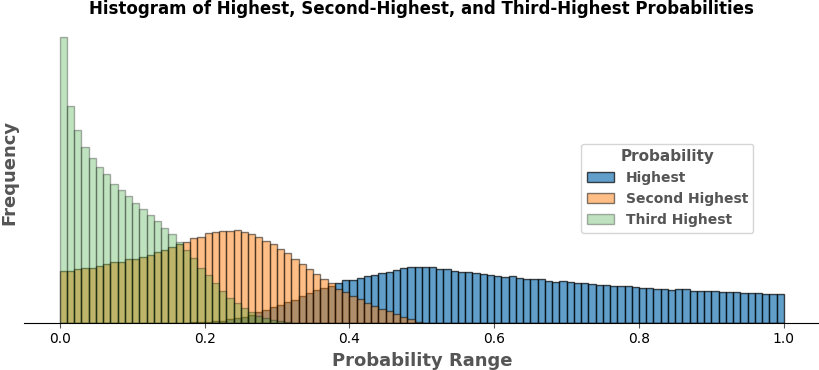}\par(c)
    \caption{(a) Standard deviation of fluctuation noises over the epochs; (b) frequency of different mineral classes having top K proportions in the mixed spectra; (c) Distribution of top 3 proportions of the combined spectra in the dataset. These distributions were observed during the training of the best architecture (IV-B) and were consistent across all architectures.}
    \label{fig:data}
\end{figure}

The training parameters are set to optimize model performance while controlling computational overhead:
\begin{itemize}
    \item \textbf{Batch Size}: The batch size is set to 50, allowing balanced updates to model weights per iteration. For the synthetic dataset, 100 steps per epoch are used to ensure robust learning with sufficient data diversity. The data generator applies data augmentation techniques to the synthetic dataset to increase generalizability.
    
    \item \textbf{Optimizer}: We use the Adam optimizer, which combines the benefits of Adaptive Gradient Algorithm (AdaGrad) and Root Mean Square Propagation (RMSProp) by adapting the learning rate based on the first and second moments of the gradients. Adam is chosen for its adaptive capability and faster convergence, given by:
    \begin{equation}
    m_t = \beta_1 m_{t-1} + (1 - \beta_1) \nabla L(\theta_{t-1})
    \end{equation}
    \begin{equation}
    v_t = \beta_2 v_{t-1} + (1 - \beta_2) \nabla L(\theta_{t-1})^2
    \end{equation}
    \begin{equation}
    \theta_t = \theta_{t-1} - \frac{\alpha}{\sqrt{v_t} + \epsilon} m_t
    \end{equation}
    where \(\alpha\) is the learning rate, \(\beta_1\) and \(\beta_2\) are decay rates, and \(\epsilon\) is a small constant.

    \item \textbf{Learning Rate and Scheduler}: The initial learning rate is set to 0.0001, with a learning rate scheduler that reduces the rate by a factor of 0.1 if validation loss does not improve for 10 consecutive epochs. This helps in refining the learning pace as training progresses, thereby preventing oscillations and local minima trapping.

    \item \textbf{Loss Function}: The mean squared error (MSE) loss is used to measure the error between the predicted and target spectra, as given by:
    \begin{equation}
    \text{MSE} = \frac{1}{n} \sum_{i=1}^{n} (y_i - \hat{y}_i)^2
    \end{equation}
    where \(y_i\) represents the target spectrum and \(\hat{y}_i\) represents the predicted spectrum. MSE is chosen for its effectiveness in penalizing large errors, thus encouraging the model to enhance preprocessing for each spectral feature.

    \item \textbf{Training data specification}: To train the proposed model, synthetic data were generated by mixing spectra from the MICA spectral library with TRDR CRISM hyperspectral data in varying proportions. The mineral with the highest proportion in the mixed spectra was labeled as the true class. Pure spectra of the true class mineral were processed using the classical preprocessing pipeline to generate the target preprocessed spectra. Gaussian noise was incrementally added to the synthetic mixed spectra to simulate real-world variability. During the initial 20 epochs, noise with a random standard deviation between 0 and 0.02 was applied. As training progressed, the noise level was gradually increased, with the final 20 epochs (81–100) incorporating noise with a random standard deviation between 0 and 0.1. This augmentation strategy ensured the model's robustness to real-world spectral fluctuations. 
    Training was conducted for up to 100 epochs for each architecture, with early stopping implemented if validation loss did not improve over 10 consecutive epochs. Each synthetic mixed spectrum is min-max scaled before processed by the model. 
\end{itemize}
Figure \ref{fig:data}(a) displays the standard deviation values of Gaussian noise added during training, grouped into five epoch segments, highlighting the progressive increase in noise intensity.
Figure \ref{fig:data}(b) shows the frequency of each mineral class appearing in the at ranks 1, 2, 3, 4, and 5 of the random proportions in the mixed spectra. The uniform distribution at each rank by each of the 28 mineral classes demonstrate the unbiasedness and consistency in the training data across classes.
Figure \ref{fig:data}(c) shows the histogram of highest, second-highest, and third-highest probabilities in the proportions exhibit bell-shaped distributions for each rank. These bell-shaped distributions show a smooth and consistent decline away from their respective peaks, indicating the gaussian variability in the proportions of mixed spectra across the ranks.

Regarding the MICAnet classification model used for evaluation, we adopt the same specifications and tuning parameters as outlined in the previous work \cite{kumari2024micanet}. This ensures consistent evaluation metrics and allows a direct comparison between the preprocessing effectiveness of the proposed model and the conventional pipeline.
\begin{table}[!t]
\centering
\caption{Ablation Study Results for Different Architectures on Synthetic and TRDR Datasets}
\label{tab:ablation}
\resizebox{\linewidth}{!}{
\begin{tabular}{ccccc}
\toprule
\textbf{\makecell[c]{Model\\Depth}} & \textbf{\makecell[c]{Encoder\\Block}} & \textbf{\makecell[c]{FLOPs\\(in Millions)}} & \textbf{\makecell[c]{Accuracy\\(Synthetic)}} & \textbf{\makecell[c]{Accuracy\\(TRDR)}} \\
\midrule
\multirow{3}{*}{I}  & A & 0.18 & 85.2\% & 82.5\% \\
 & B & 3.12 & 86.3\% & 83.0\% \\
 & C & 6.11 & 85.9\% & 82.8\% \\
\midrule
\multirow{3}{*}{II}   & A & 9.07 & 88.1\% & 85.0\% \\
 & B & 16.45 & 89.2\% & 86.0\% \\
 & C & 20.93 & 88.5\% & 85.6\% \\
\midrule
\multirow{3}{*}{III} & A & 13.51 & 89.5\% & 87.2\% \\
 & B & 23.11 & 90.3\% & 88.1\% \\
 & C & 28.34 & 89.8\% & 87.5\% \\
\midrule
\multirow{3}{*}{IV} & A & 15.74 & 90.1\% & 87.8\% \\
 & B & 26.44 & 91.0\% & 89.0\% \\
 & C & 32.04 & 90.5\% & 88.5\% \\
\bottomrule
\end{tabular}
}
\end{table}
\paragraph{Ablation Study}\label{sec:ablationstudy}
To determine the optimal configuration for our 1D-UNet preprocessing model, we conducted an ablation study using various architectural setups. Each configuration consists of different numbers of encoder and decoder blocks, denoted as $N$, as well as various arrangements within each encoder block. The primary aim of this study is to assess the trade-off between computational complexity and accuracy achieved on both synthetic and TRDR datasets when using the MICAnet classification model to evaluate preprocessing effectiveness.
The experimental setups are divided into four main architectures, each with an increasing value of $N$:
\begin{itemize}
    \item \textbf{I.} $N=0 :$ Baseline architecture with only the bottleneck block.
    \item \textbf{II.} $N=1 :$ Intermediate architecture with 1 decoder and 1 encoder blocks.
    \item \textbf{III.} $N=2 :$ Proposed architecture with 2 decoder and 2 encoder blocks.
    \item \textbf{IV.} $N=3 :$ Deeper architecture with 3 decoder and 3 encoder blocks.
\end{itemize}

For each of these main architectures (I-IV), we tested three variations of the encoder block configuration:
\begin{itemize}
    \item \textbf{A.} 1 convolutional layer (stride = 1) followed by 1 max-pooling layer (stride = 2).
    \item \textbf{B.} 1 convolutional layer (stride = 1) followed by 1 convolutional layer (stride = 2).
    \item \textbf{C.} 2 convolutional layers (stride = 1) followed by 1 max-pooling layer (stride = 2).
\end{itemize}

The performance of each setup is measured in terms of floating-point operations (FLOPs) and accuracy, where the accuracy is obtained using the MICAnet classifier on both the synthetic dataset and the CRISM TRDR dataset. The results of this ablation study are summarized in Table~\ref{tab:ablation}.

The results presented in Table~\ref{tab:ablation} reveal the impact of different architectural configurations on both computational complexity and classification accuracy. We observe the following trends:
\begin{enumerate}
    \item \textbf{Increasing Depth Improves Accuracy:} As we move from Architecture I to Architecture IV (increasing the number of encoder-decoder pairs), there is a consistent improvement in accuracy across both datasets. Architecture IV-B, with three encoder-decoder pairs and the B configuration in each encoder block, achieves the highest accuracy of 91.0\% on the synthetic dataset and 89.0\% on the TRDR dataset. This indicates that a deeper network can better capture the spectral features necessary for mineral identification.

    \item \textbf{Trade-Off Between FLOPs and Accuracy:} There is a noticeable increase in FLOPs with deeper architectures and more complex encoder block configurations (particularly in B and C). For instance, Architecture I-A requires only 0.18 million FLOPs, whereas Architecture IV-B requires 26.44 million FLOPs. Despite the computational cost, the accuracy improvements suggest that the increased complexity enhances feature extraction and contributes to better overall performance.

    \item \textbf{Encoder Block Configuration Impact:} Among the three configurations within each architecture, Configuration B (one convolutional layer with stride 1, followed by one convolution layer with stride 2 and no max-pool layers) tends to yield the highest accuracy. This setup likely enhances the model’s ability to capture subtle spectral variations, which are essential for distinguishing minerals. Configuration C (with two convolutional layers of stride 1 followed by a max-pooling layer of stride 2) achieves moderately high accuracy but slightly lower than B due to the potentially reduced spatial information retention.
\end{enumerate}

The optimal setup for our 1D-UNet preprocessing model is Architecture IV-B, which offers a balanced trade-off between computational complexity and accuracy. With three encoder-decoder pairs and each encoder block containing two convolutional layers (stride 1) followed by an upscaling layer, this configuration achieves high accuracy while retaining significant spectral information necessary for mineral identification.

\begin{figure}[!b]
    \centering
    \includegraphics[width=\linewidth]{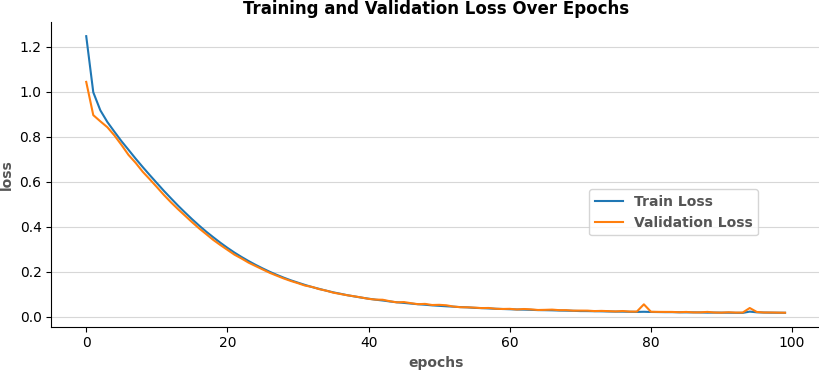}
    \caption{Training and validation loss over epochs during the training of the best architecture IV-B.}
    \label{fig:loss}
\end{figure}
\begin{figure*}[!t]
    \centering
    \includegraphics[width=\linewidth]{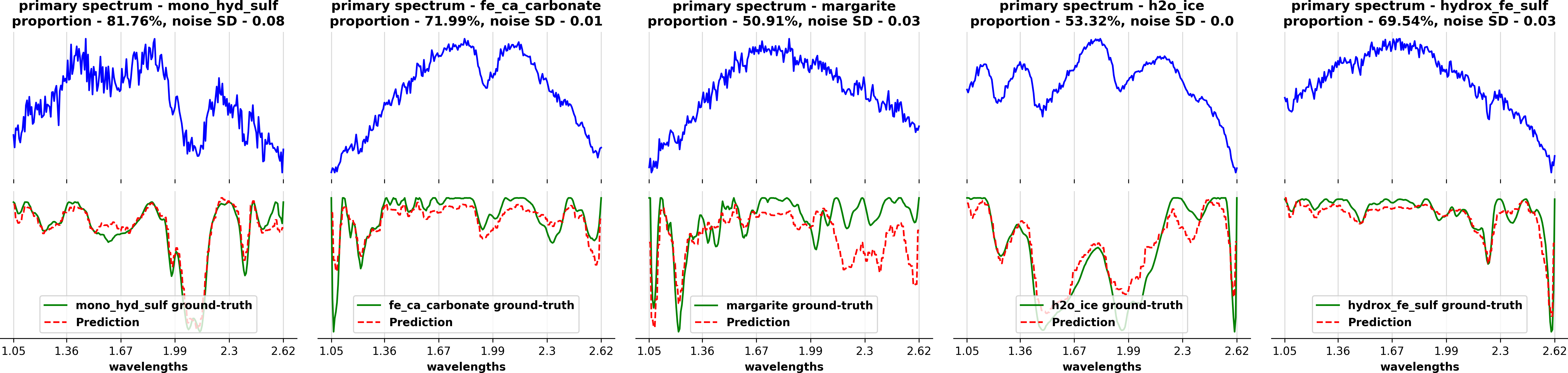}
    \caption{Some sample synthetic spectra, and their preprocessing results by architecture IV-B compared to the ground truth.}
    \label{fig:samples}
\end{figure*}
\subsection{Performance Analysis and Comparison with Traditional Preprocessing Pipeline}
The training and validation loss curves for the best architecture (IV-B), in figure \ref{fig:loss}, demonstrate a smooth convergence over 100 epochs. Both loss curves start high at the beginning of training but rapidly decrease during the initial epochs, indicating effective learning. Beyond approximately 40 epochs, the losses stabilize together and approach near-zero values, reflecting minimal overfitting and strong generalization. The close overlap between the training and validation loss curves highlights the model's robustness, as it maintains consistent performance on both the training and validation datasets. Minor fluctuations observed towards the later epochs are likely due to the progressive addition of noise during training.

\begin{figure}[!t]
    \centering
    \includegraphics[width=\linewidth]{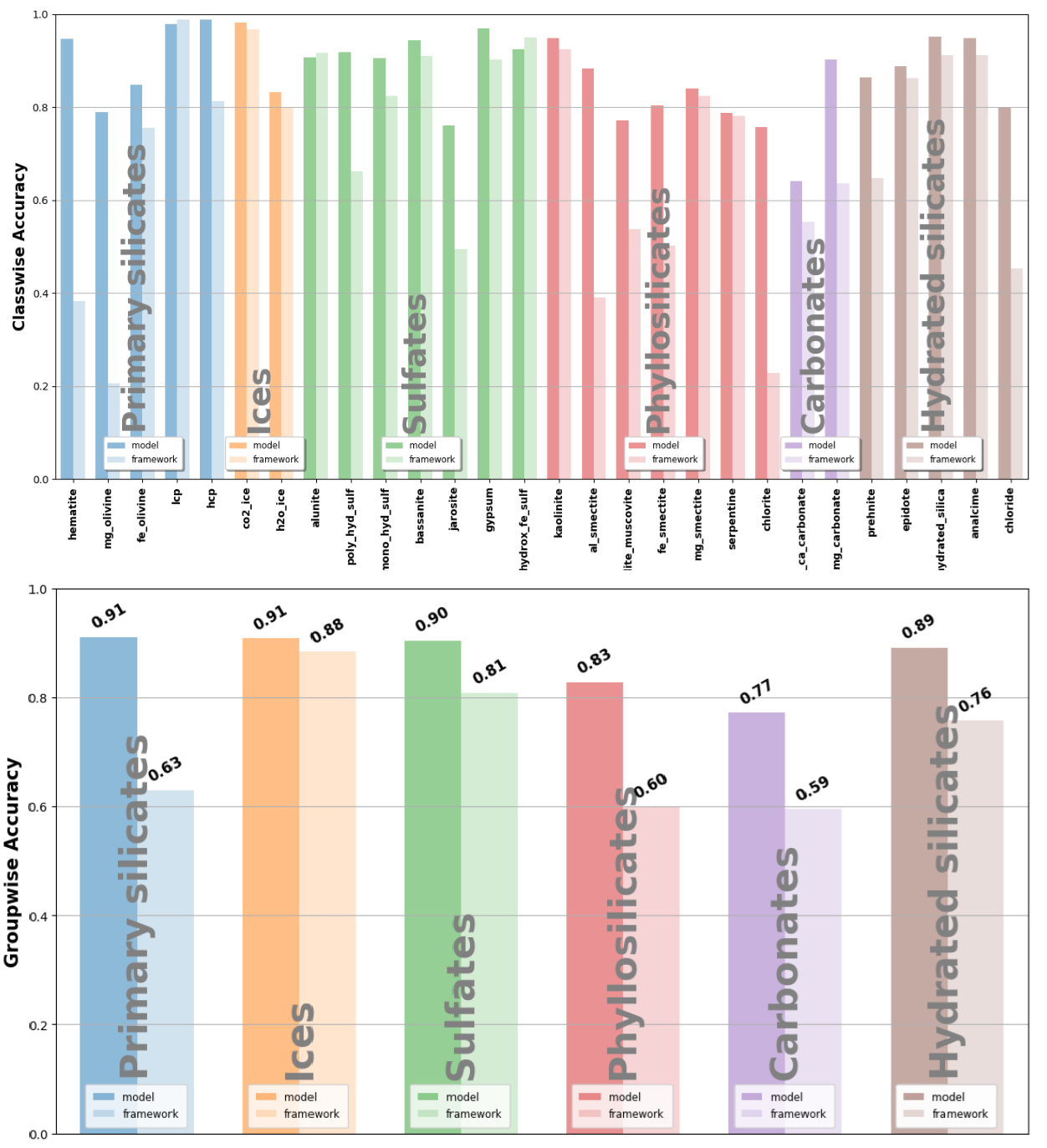}
    \caption{Classwise and group-wise accuracy of the minerals considered in the experiment shows detection improvement of around 5\% if the proposed preprocessing model is used than the framework.}
    \label{fig:comparison}
\end{figure}

Figure \ref{fig:samples} shows the preprocessed spectra of some synthetic mineral spectra by architecture IV-B, including kaolinite, epidote, analcime, and hematite. The preprocessed spectra are very close to the ground-truth preprocessed spectra (primary mineral spectra passed through the classical preprocessing pipeline) even when the proportions of the primary minerals are significantly less.
The performance of the architecture IV-B is highlighted in detail in figure \ref{fig:comparison}. Training was performed using augmented spectra from the MICA spectral library, while testing employed TRDR CRISM hyperspectral data \cite{plebani2022machine}. The classwise and groupwise accuracy results show the model's robustness and efficiency. For individual mineral classes, the proposed framework consistently outperformed the classical pipeline, achieving higher accuracies across 28 mineral classes. Similarly, in groupwise evaluation, the proposed model achieved notable improvements, particularly for primary silicates, ices, and sulfates, with average groupwise accuracies of 0.91, 0.91, and 0.90, respectively, and also there is a significant enhancement in the accuracy of the carbonate minerals. 
%
%
% %% %% %% %% %% %% %% %% %% %% %% %% %% %% %% %% %% %% %% %% %% %% %% %%
% %% %% %%   P E R F O R M A N C E   A N A L Y S I S    %% %% %% %% %% %% 
% %% %% %% %% %% %% %% %% %% %% %% %% %% %% %% %% %% %% %% %% %% %% %% %%
%
%
\section{Mineral Mapping on the Martian Surface}\label{sec:expresults}
    
\begin{figure}[!t]
    \centering
    \includegraphics[width=\linewidth]{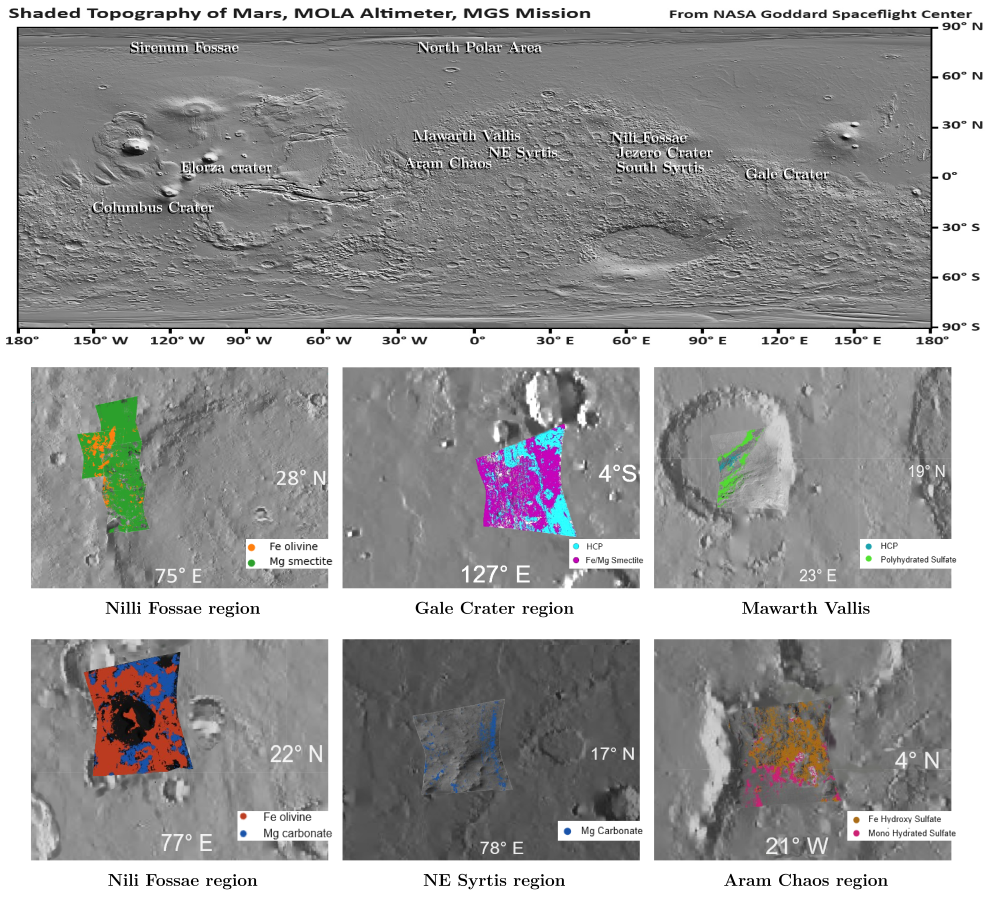}
    \caption{Top: Different locations on Martian surface marked on Mola elevation map. Bottom: Mineral mapping on different MTRDR data super imposed on Mola elevation map. Note that, The resolution of the elevation map and the hyperspectral data are not the same.}
    \label{fig:result}
\end{figure}

The Syrtis Major quadrangle on Mars, which spans latitudes $0$° to $30$° north and longitudes $270$° to $315$° west, features a significant area of lowlands known as the Nili Fossae. This region is primarily composed of olivines, smectites, hydrated silica, kaolinite, and iron oxides \cite{hoefen2003discovery,edwards2015carbon}. In September 2015, Nili Fossae was selected as a potential landing site for the Mars 2020 rover. Jezero Crater, an extensive impact structure with a diameter of approximately $45$ kilometers \cite{29}, is located within this area and served as the landing site for the Mars 2020 mission. This mission aims to collect mineral samples that may eventually be returned to Earth by future expeditions, allowing for a comparison of mineral diversity over time within this specific region of the Martian surface.
Mawrth Vallis is another significant geological feature, measuring $100$ kilometers across, which has been shaped by various impact events, small streams, and volcanic activity. Studies utilizing visible and near-infrared spectrometers have extensively examined Mawrth Vallis, identifying key minerals such as Fe/Mg-phyllosilicates, Al-phyllosilicates, and high-calcium pyroxene (HCP) \cite{34,35}.
Northeast Syrtis, part of the Syrtis Major volcanic province located in the northern hemisphere of Mars, showcases stratified terrain rich in a diverse array of igneous minerals, including olivine and both high- and low-calcium pyroxenes. Additionally, aqueous minerals such as clays, carbonates, serpentine, and sulfates are also prevalent in this area \cite{murchie2009synthesis,ehlmann2009identification}.
Located in the Margaritifer Terra region, the Aram Chaos impact crater spans $280$ kilometers in diameter. This region is known for its frequent occurrences of minerals such as hematite, jarosite, and hydrated sulfates \cite{glotch2005geologic}.
Figure \ref{fig:result} shows the locations of the regions on the Martian surface .

The minerals identified on the Martian surface exhibit a diverse array of diagnostic absorption features in the VNIR spectral range of 1 to 2.6 micrometers. Hydrated silicates, similar to Fe/Mg-smectites, show a combination of features, with a broad absorption around 1.4 micrometers due to OH stretching, a band near 1.9 micrometers from H\textsubscript{2}O bending, and a feature around 2.2-2.3 micrometers associated with Al-OH and Mg-OH vibrations. Sulfates, such as gypsum, exhibit a suite of diagnostic features, including bands at 1.45, 1.75, 1.94, 2.22, and 2.48 micrometers, arising from combinations of H\textsubscript{2}O stretching and bending, as well as SiO\textsubscript{4} stretching vibrations. Additionally, Fe-hydroxy sulfates and mono/poly-hydrated sulfates exhibit distinct absorptions related to OH, H\textsubscript{2}O, and SiO\textsubscript{4} vibrational modes in this spectral region.

The discovery of minerals such as Mg-carbonate, Fe/Mg smectite, HCP, and Mg-Olivine CRISM hyperspectral image from Mars has significant implications for understanding the geological processes, environmental conditions, and potential resource exploration opportunities in the studied area. The presence of Mg-carbonate suggests past aqueous alteration processes, as carbonates typically form in the presence of liquid water and carbon dioxide. Fe/Mg smectites are clay minerals that form through the alteration of igneous rocks in the presence of water, providing evidence of past water-rock interactions. The formation of Mg-carbonate and Fe/Mg smectites imply the existence of past habitable environments with the presence of liquid water, potentially for extended periods. The presence of Mg-carbonate and Fe/Mg smectites also could indicate the existence of significant subsurface water reservoirs, which could be valuable resources for future human exploration and settlement on Mars. The stability of these minerals can provide constraints on the pH, temperature, and water chemistry conditions that existed during their formation. Mg-olivine and HCP are primarily igneous minerals, indicating the presence of mafic or ultramafic rocks, which can provide insights into the region’s volcanic and magmatic history. Mg-olivine is a potential source of magnesium, which could be extracted and utilized for various purposes, such as construction materials or soil enhancement for future Martian agriculture. The identification of these minerals can help prioritize landing sites and regions of interest for future robotic and human exploration missions, as well as for resource prospecting and utilization.

The detection of Fe-olivine and Mg-carbonate in the Nili Fossae region (FRT3E12)
indicates the presence of mafic or ultramafic igneous rocks, providing insights into
the region’s volcanic and magmatic history, as well as potential aqueous alteration
processes. The
presence of Fe-hydroxy sulfate and mono-hydrated sulfate in the Aram Chaos region
(FRT98B2) also points to the existence of sulfate-rich deposits, potentially formed
through the alteration of sulfide-bearing rocks or volcanic activity. The detection of
Mg-carbonate in the NE Syrtis region (FRT19538) further supports the hypothesis of
past aqueous alteration processes and the potential for habitable environments in this
area. The identification of high-calcium pyroxene and poly-hydrated sulfate in the Gale
Crater region (FRTC518) suggests the presence of mafic igneous rocks and sulfate-rich
deposits, respectively, providing insights into the region’s geological history and potential
past environments. The presence of high-calcium pyroxene and Fe/Mg-smectites in
the Mawrth Vallis region (FRT9326) indicates the presence of mafic igneous rocks and
clay minerals, suggesting water-rock interactions and potential habitable environments
in this region. These findings highly contribute to our understanding of the geological composition,
evolution, and potential past habitable environments in these regions, which
is crucial for unraveling Mars’ geological and potential biological history.
%
%
% %% %% %% %% %% %% %% %% %% %% %% %% %% %% %% %% %% %% %% %% %% %% %% %%
% %% %% %% %% %% %%  C O N C L U S I O N S  %% %% %% %% %% %% %% %% %% %%
% %% %% %% %% %% %% %% %% %% %% %% %% %% %% %% %% %% %% %% %% %% %% %% %%
%
%
\section{Discussion and Conclusion}\label{sec:conclusion}

The results of this study demonstrate the efficacy of the proposed UNet-based preprocessing model for CRISM MTRDR hyperspectral data in identifying minerals on the Martian surface. By integrating the preprocessing steps typically performed in traditional pipelines into a unified neural network framework, we significantly reduce the time and computational resources required for mineral classification. The model not only streamlines the preprocessing phase but also enhances the quality of the spectral data, improving the distinguishability of mineral absorption features.

The augmented training dataset created from the MICA spectral library, comprising 31 mineral spectra across six mineral groups, has proven to be a crucial component in training the model effectively. This approach mitigates the challenge of limited labeled data for Martian mineral identification, allowing for a more robust learning process. The ability to randomly sample and proportion minerals within the training dataset enables the model to generalize better, accommodating the diverse mineral mixtures that are frequently encountered in CRISM hyperspectral data.

The performance analysis highlights a marked improvement in both accuracy and efficiency when using the proposed model compared to traditional preprocessing pipelines. The findings suggest that the model maintains high classification accuracy while significantly reducing processing time, making it suitable for real-time applications in planetary exploration.
Furthermore, the potential of the model extends beyond mineral identification on Mars. The framework could be adapted for similar applications in other planetary bodies or even for terrestrial remote sensing data. Future research could explore the integration of additional features, such as temporal data, to further enhance mineral identification capabilities.

\section*{Acknowledgement}
We thank the anonymous reviewers for their constructive feedback, which helped improve the quality of this paper. We also gratefully acknowledge the support of The ISPRS Foundation (TIF) for awarding the travel grant to attend the ISPRS Geospatial Week 2025 in Dubai, which provided a valuable opportunity to present this research and gain insights from the global scientific community. 

%
%
% %% %% %% %% %% %% %% %% %% %% %% %% %% %% %% %% %% %% %% %% %% %% %% %%
% %% %% %% %% %% %%   R E F E R E N C E S   %% %% %% %% %% %% %% %% %% %%
% %% %% %% %% %% %% %% %% %% %% %% %% %% %% %% %% %% %% %% %% %% %% %% %%
%
%
{
	\begin{spacing}{1.17}
		\normalsize
		\bibliography{main} % Include your own bibliography (*.bib), style is given in isprs.cls
	\end{spacing}
}

\end{document}